%% file: paper.tex
\documentclass{article}
\usepackage{mysty}

\usepackage{times}
\usepackage{soul}
\usepackage{url}
\usepackage[hidelinks]{hyperref}
\usepackage[utf8]{inputenc}
\usepackage[small]{caption}
\usepackage{graphicx}
\usepackage{amsmath}
\usepackage{amsthm}
\usepackage{booktabs}
\usepackage{algorithm}
\usepackage{algorithmic}
\usepackage{multirow}
\usepackage{amssymb}
\usepackage{enumitem}

\input{preamble}

\title{On Modifying a Neural Network's Perception}

\author{
    Manuel de Sousa Ribeiro
    \and
    Jo\~ao Leite
	\affiliations
    NOVA LINCS, NOVA University Lisbon, Portugal
    \emails
    mad.ribeiro@campus.fct.unl.pt,
    jleite@fct.unl.pt
}

\begin{document}

\maketitle

\input{abstract}

\input{introduction}

\input{method}
\input{identifying_neurons}

\input{manipulating_activations}

\input{importance_of_neurons}

\input{methods_cost}

\input{generating_counterfactuals}

\input{correcting_errors}

\input{validation_realworld_data}

\input{related_work}

\input{conclusion}

\bibliographystyle{named}
\bibliography{biblio}

\end{document}

%% file: preamble.tex
\newcommand{\nna}{$\mathcal{M}_\mathsf{A}$}

\newcommand{\concept}[1]{\mathsf{#1}} 

\newcommand{\reals}{\mathbb{R}}

\newcommand{\inp}{\mathbb{I}} 
\newcommand{\out}{\mathbb{O}}

%% file: abstract.tex
\begin{abstract}
	Artificial neural networks have proven to be extremely useful models that have allowed for multiple recent breakthroughs in the field of Artificial Intelligence and many others. However, they are typically regarded as \emph{black boxes}, given how difficult it is for humans to interpret how these models reach their results.
	In this work, we propose a method which allows one to modify what an artificial neural network is perceiving regarding specific human-defined concepts, enabling the generation of hypothetical scenarios that could help understand and even debug the neural network model.
	Through empirical evaluation, in a synthetic dataset and in the ImageNet dataset, we test the proposed method on different models, assessing whether the performed manipulations are well interpreted by the models, and analyzing how they react to them.
\end{abstract}

%% file: introduction.tex
\section{Introduction}

In this paper, we investigate how to modify a neural network's perception regarding specific human-defined concepts not directly encoded in its input, with the goal of being able to better understand how such concepts affect a models' predictions.
Our results suggest that this is possible to do with few labeled data, and without need to change or retrain the existing neural network model.

In the last few decades, artificial neural networks have enabled major advances in multiple fields, from text, image, video and speech processing \cite{Khan2020}, to medicine \cite{Shahid2019}, economics and finance \cite{Ghoddusi2019}, and many others. Numerous successful neural network-based applications have recently been implemented \cite{Abiodun2018}, rendering these models ubiquitous.

Despite their popularity, neural networks lack interpretability, as they supply no direct human-interpretable indication of why a given result was provided \cite{Hitzler2020}. This led to the development of multiple methods focused on solving this shortcoming (cf. Section \ref{sec:related_work}). Most popular methods typically focus on pointing out which inputs contributed most for the output, or on substituting the model for one that is inherently interpretable \cite{Kim2018}.
However, such methods leave to the users the burden of understanding why the provided explanation justifies the output, e.g, users must interpret why a particular set of features, e.g., pixels in an image, and their values, leads to the output.
Various user studies \cite{Adebayo2020,Chu2020,Shen2020} show that the explanations given by these methods are often disregarded or unhelpful to end users.

One reason is that humans do not typically reason with features at a very low level, such as individual pixels in an image -- they typically reason with higher-level human defined concepts. For example, a useful explanation from the standpoint of a human as to why a network classified a particular picture of a train as being one of a passenger train will probably refer to the fact that the wagons had windows rather than pointing out specific pixels in the image.

Additionally, when humans attempt to determine the causes for some non-trivial phenomena, they often resort to counterfactual reasoning \cite{Miller2019}, trying to understand how different scenarios would lead to different outcomes. 
This approach seems also helpful for interpreting artificial neural networks, as it emphasizes the causes -- what is changed -- and the effects -- what mutates as a result of the changes.
For example, to better interpret how a neural networks is classifying a particular picture of a train, the user could ask what would have been the output had the picture contained a passenger wagon.
We would like to be able to generate such counterfactual scenarios based on how different human-defined concepts impact a model's predictions. The use of human-defined concepts -- the concept of passenger wagon in the previous example -- is important as it provides semantics with which to describe the counterfactuals, while ensuring that these concepts are meaningful and understandable for the users of a particular model.

There has been work on developing methods to generate counterfactuals for artificial neural networks \cite{Guidotti2022}, with some even allowing for generating counterfactual samples with respect to particular attributes \cite{Yang2021}. However, to our knowledge, they focus on how particular changes to the input samples might affect a models' output, neglecting what the model is actually perceiving from those samples.
Furthermore, current methods to produce counterfactuals are typically complex to implement, often requiring the training of an additional model to produce the counterfactuals \cite{Stepin2021}, and the use of specific neural network architectures, e.g., invertible neural networks \cite{Hvilshoej2021}.

In this work, we address the issue of counterfactual generation at a different level of abstraction, focusing on what a model is perceiving from a given input regarding specific human-defined concepts, and how that affects the model's output. The idea would be to ``\emph{convince}'' the model that it is perceiving a particular concept -- for example a passenger wagon -- without producing a specific image containing one, and checking its effect on the model's output.
By abstracting away from generating particular counterfactual samples and instead focusing on generating counterfactuals with regards to what a model is perceiving about human-defined concepts, we allow for a better understanding of how what is encoded in a neural network impacts a models' predictions in a human-understandable manner.
By manipulating what a model perceives regarding specific concepts of interest, we allow users to explore how the output of a neural network depends on what it is perceiving regarding such concepts.

In order for it to be possible to generate counterfactuals based on what a model is perceiving instead of what the model is fed, we need to be able to understand what a neural network model is perceiving and be able to manipulate what the model is perceiving with respect to a specific concept.
We find inspiration in the research conducted in the field of neuroscience, where highly selective neurons that seemed to represent the meaning of a specific stimulus can be identified \cite{Reddy2014}. These neurons, known as \emph{concept cells}, seemed to ``provide a sparse, explicit and invariant representation of concepts'' \cite{Quiroga2005,Quiroga2012}. The discovery of these cells contributed for a better understanding of how the brain -- a complex and subsymbolic system -- works and how it relates different concepts \cite{Gastaldi2022}.
Additionally, through the technique of optogenetics, it is possible to manipulate the activity of specific neurons and learn their purpose \cite{Okuyama2018}.

Analogously, being able to assign meaning to specific neurons in a neural network could provide us with a better understanding of what information is encoded in a given model, and how it might be associating different concepts.
Moreover, given that typically one has access to a neural networks' internals, we could manipulate the outputs of the neurons to which meaning has been assigned, and examine how such changes affect the model under different circumstances, thus generating counterfactual scenarios.

Based on the existing evidence that neurons with similar properties to concept cells seem to emerge in artificial neural networks \cite{Goh2021}, we hypothesize that by identifying which neurons in a neural network act as concept cells to specific human-defined concepts and by manipulating the activations of such neurons, we should be able to modify a neural network's perception regarding those concepts. 

In this paper, we propose and test a method to generate counterfactuals scenarios for artificial neural networks by modifying what they are perceiving regarding specific human-defined concepts.
In Section \ref{sec:method}, we present the proposed method, with Section \ref{sec:identify_neurons} discussing how to pinpoint which neurons identify a particular concept in a neural network, and Sections \ref{sec:manipulate_perception} to \ref{sec:cost} addressing different aspects of the proposed method and providing experimental evidence to support our claims. Sections \ref{sec:interpret_nns} and \ref{sec:correct_nns} illustrate possible applications of the proposed method. In Section \ref{sec:validation}, we apply and test the method in the setting of the ImageNet dataset. In Section \ref{sec:related_work}, we discuss related work, concluding in Section \ref{sec:conclusions}.

%% file: method.tex
\section{A Method to Manipulate a Neural Network's Perception}
\label{sec:method}

In order to generate counterfactuals regarding what a neural network model is perceiving about human-defined concepts, we propose a method composed by three main steps. For each concept of interest:
\begin{enumerate}[label=\alph*)]
	\item Estimate how sensitive each neuron is to that concept, i.e., how well its activations separate samples where that concept is present from those where it is absent;
	\item Based on the neurons' sensitivity values, select which neurons are considered as ``concept cell-like'', which we will refer to as \emph{concept neurons};
	\item For each concept neuron, compute two activation values, representing, respectively, the output of that neuron for samples where that concept is present and absent.
\end{enumerate}

Consider a neural network model $\mathcal{M} : \inp \rightarrow \out$, which is the model being analyzed, where $\inp$ is the input data space and $\out$ denotes the model output space. For each considered human-defined concept $\concept{C}$, we assume a set of positive $P_{\concept{C}} \subseteq \inp$ and negative $N_{\concept{C}}  \subseteq \inp$ samples where the concept is, respectively, present/absent. Let $a_{i}^{\mathcal{M}}: \inp \rightarrow \reals$ represent the output of the $i$\textsuperscript{th} neuron of neural network model $\mathcal{M}$ according to some arbitrary ordering of the neurons.

The first step of our method consists in estimating how sensitive each neuron is to each considered concept, i.e., how well the activations of each neuron separate samples where a concept is present from those where it is not.
We denote by $r_{i}^{\mathcal{M}} : 2^\inp \times 2^\inp \rightarrow [0, 1]$ the function which, for a given neuron $i$ of a model $\mathcal{M}$, takes a set $P_{\concept{C}}$ and $N_{\concept{C}}$ and provides a value representing how sensitive $i$ is to concept $\concept{C}$.
In Section \ref{sec:identify_neurons}, we consider different implementations of this function.

\input{ontology}

The second step is to select, for each concept of interest $\concept{C}$, which neurons of $\mathcal{M}$ are to be considered as concept neurons for $\concept{C}$. Let $s_{\concept{C}} : \reals \rightarrow \{0,1\}$ be a threshold function indicating, for a given concept $C$ and sensitivity value, whether that value is high enough to be considered as a concept cell.
Then, for each concept of interest $\concept{C}$, we determine the set of selected neurons as $S_{\concept{C}} = \{i : s_{\concept{C}}(r_{i}^{\mathcal{M}}(P_{\concept{C}}, N_{\concept{C}})) = 1, i \in [1, k]\}$, where $k$ is the number of neurons in $\mathcal{M}$.
In all experiments, we set the threshold value by gradually decreasing it, while testing the model performance in a $100$ sample validation set for each concept. Once more than $3$ neurons have been added, by decreasing the threshold, and no performance improvement is seen in the validation set, we set the threshold to the best found value.

Lastly, for each neuron selected as a concept neuron for some concept $\concept{C}$, we compute two activation values representing, respectively, when that concept is present and absent. We consider a function $c_{i}^{\mathcal{M}} : 2^\inp \rightarrow \reals$, which computes an activation value for the $i$\textsuperscript{th} neuron of $\mathcal{M}$, and use it compute an activation value representing the presence of concept $\concept{C}$, $c_{i}^{\mathcal{M}}(P_{\concept{C}})$, and one representing its absence, $c_{i}^{\mathcal{M}}(N_{\concept{C}})$. 
In Section \ref{sec:manipulate_perception}, we discuss different implementations of $c_{i}^{\mathcal{M}}$ and compare how they impact the method's results.

Subsequently, when some input $x \in \inp$ is fed to neural network $\mathcal{M}$, to generate a counterfactual scenario where concept $\concept{C}$ is present or absent, we only need to replace the activation value $a_{i}^{\mathcal{M}}(x)$ of each neuron $i$ in $S_{\concept{C}}$ with $c_{i}^{\mathcal{M}}(P_{\concept{C}})$ or $c_{i}^{\mathcal{M}}(N_{\concept{C}})$, respectively. We refer to this step, as \emph{injecting} a concept into a neural network model.

\begin{figure}
	\hfill
	\hfill
	\includegraphics[width=0.10\textwidth]{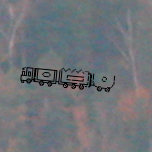}
	\hfill
	\includegraphics[width=0.10\textwidth]{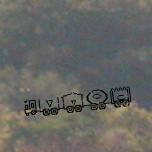}	
	\hfill
	\includegraphics[width=0.10\textwidth]{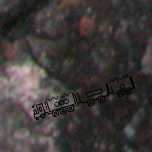}	
	\hfill
	\hfill
	\caption{Sample images of the XTRAINS dataset.}
	\label{fig:xtrains_samples}
\end{figure}

To test our method, we consider the Explainable Abstract Trains Dataset (XTRAINS) \cite{SousaRibeiro2020}, a synthetic dataset composed of representations of trains, such as those shown in Figure \ref{fig:xtrains_samples}. This dataset contains labels regarding various visual concepts and a logic-based ontology describing how each of these concepts is related.
Figure \ref{fig:xtrains_ontology} shows a subset of the dataset's accompanying ontology illustrating how different concepts are related to each other. This ontology specifies, for example, that $\sf TypeA$ is either $\sf WarTrain$ or $\sf EmptyTrain$, and that $\sf WarTrain$ encompasses those having a $\sf ReinforcedCar$ and a $\sf PassengerCar$. These concepts have a visual representation, e.g, a $\sf ReinforcedCar$ is shown as a car having two lines on each wall, such as the first two cars of the leftmost image in Figure \ref{fig:xtrains_samples}.

In the experiments described below, we adopt a neural network developed in \cite{Ferreira2022}, trained to identify trains of $\sf TypeA$ -- referred to as \nna. This neural network was trained to achieve an accuracy of about $99\%$ in a balanced test set of $10\ 000$ images.
We also consider the definition of relevancy given in \cite{SousaRibeiro2021} to establish which concepts are related to the task of a given neural network (w.r.t. the dataset's accompanying ontology).

%% file: ontology.tex
\begin{figure*}
	\setlength{\multlinegap}{0pt}
	\setlength{\abovedisplayskip}{0pt}
	\setlength{\belowdisplayskip}{0pt}
	\setlength{\abovedisplayshortskip}{0pt}
	\setlength{\belowdisplayshortskip}{0pt}
	\setlength{\jot}{0pt}
	\fontsize{8.0pt}{8.0pt} \selectfont
	\begin{flalign*}
		& \sf TypeA \equiv WarTrain \sqcup EmptyTrain & \exists \sf has.FreightWagon \sqcap \exists has.PassengerCar \sqcap \exists has.EmptyWagon & \sf \sqsubseteq MixedTrain \\
		& \sf TypeB \equiv PassengerTrain \sqcup LongFreightTrain & \exists \sf has.(PassengerCar \sqcap LongWagon) \sqcup (\geq 2\ has.PassengerCar) & \sf \sqsubseteq PassengerTrain \\
		& \sf TypeC \equiv RuralTrain \sqcup MixedTrain & \sf \exists has.ReinforcedCar \sqcap \exists has.PassengerCar & \sf \sqsubseteq WarTrain \\
		& \sf LongFreightTrain \equiv LongTrain \sqcap FreightTrain & \sf  (\geq 2\ has.LongWagon) \sqcup (\geq 3\ has.Wagon) & \sf  \sqsubseteq LongTrain \\
		& \sf EmptyTrain \equiv \forall has.(EmptyWagon \sqcup  Locomotive) \sqcap \exists has.EmptyWagon \span \sf  (\geq 2\ has.FreightWagon) & \sf  \sqsubseteq FreightTrain
	\end{flalign*}
	\caption{Subset of the XTRAINS dataset ontology’s axioms, describing how the trains’ representations are classified.}
	\label{fig:xtrains_ontology}
\end{figure*}

%% file: identifying_neurons.tex
\subsection{Identifying \emph{Concept Cell-like} Neurons} \label{sec:identify_neurons}

In order to be able to modify a neural network's perception regarding specific human-defined concepts, it is first necessary to determine which neurons in a model are identifying these concepts. 
Based on the evidence that neural networks seem to have information encoded in their internals regarding concepts which are related with their tasks \cite{SousaRibeiro2021}, and that neurons with ``concept cell-like'' properties seems to emerge in artificial neural networks \cite{Goh2021}, we hypothesize that neural networks have neurons which act as concept cells for concepts related with the tasks they perform.
In this section, we investigate how to determine such neurons in a neural network model.

\begin{figure}
	\includegraphics[width=0.98\linewidth]{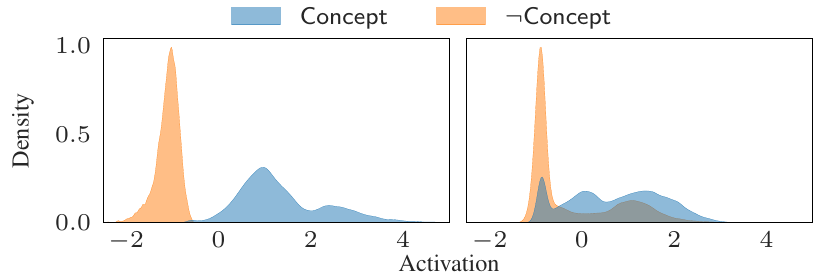}
	\caption{Probability density function of two neurons for samples where a given concept is present/absent.}
	\label{fig:act_dist}
\end{figure}

We consider a neuron to be ``concept-cell like'' for some concept $\concept{C}$ if it is possible to separate samples where $\concept{C}$ is present from those where it is absent based on its activations. 
Figure \ref{fig:act_dist}, illustrates the estimated probability density function of two neurons for some $P_{\concept{C}}$ and $N_{\concept{C}}$ sets. The neuron on the left side seems to act as a concept cell for $\concept{C}$, showing well separated distributions for both sample sets. On the other hand, the neuron on the right side does not act as a concept cell for this concept, given that its activations do not distinguish between both sets of samples.

While there are methods, such as TCAV \cite{Kim2018} and Mapping Networks \cite{SousaRibeiro2021}, which allow for an understanding of whether a given model is sensitive to a certain concept, here we are interested in understanding whether individual neurons are sensitive to that concept. Our goal is not to check whether the model is sensitive to a concept, for which only some neurons might be sufficient, but rather to find all neurons that are sensitive to that concept, so that we can manipulate them. 

We consider three different implementations of $r_{i}^{\mathcal{M}}$ to evaluate the adequacy of a neuron $i$ of $\mathcal{M}$ as concept cell for $\concept{C}$:
\begin{itemize}
	\item Spearman rank-order correlation between a neuron's activations and the dataset's labels, computed as $|1 - \frac{6\sum d_{j}^{2}}{n(n^{2}-1)}|$, where $d_{j} = a_{i}^{\mathcal{M}}({P_{\concept{C}}}_j) -  a_{i}^{\mathcal{M}}({N_{\concept{C}}}_j)$ and $n=|P_{\concept{C}}|$, assumes $|P_{\concept{C}}| = |N_{\concept{C}}|$;
	\item Accuracy of a linear binary classifier ($b$) predicting the dataset's labels from a neuron's activations, computed as $\frac{| \{x : x \in N_{\concept{C}} \cap b(a_{i}^{\mathcal{M}}(x)) = 0) \} |+ | \{x : x \in P_{\concept{C}} \cap b(a_{i}^{\mathcal{M}}(x)) = 1) \} |}{|N_{\concept{C}}|+|P_{\concept{C}}|}$;
	\item Probability density function intersection of a neuron's activations for $P_{\concept{C}}$ and $N_{\concept{C}}$, computed as $1 - \int min(f_i^{P_{\concept{C}}}(x), f_i^{N_{\concept{C}}}(x))\,dx$ where $f_i^{P_{\concept{C}}}$ is the estimated probability density function of the activation value of neuron $i$ for the positive samples in $P_{\concept{C}}$ and $f_i^{N_{\concept{C}}}$ for the negative samples, assumes the samples to be independent and identically distributed.
\end{itemize}

\input{tables/selected_neurons_nn_a}

To test our hypothesis that neural networks should typically have concept neurons for concepts which are relevant for their tasks, we compute the three described metrics for four random relevant concepts: $\concept{EmptyTrain}$, $\concept{\exists has.PassengerCar}$, $\concept{\exists has.ReinforcedCar}$, and $\concept{WarTrain}$; and for four non-relevant concepts: $\concept{\exists has.LongWagon}$, $\concept{\exists has.OpenRoofCar}$, $\concept{LongFreightTrain}$, and $\concept{MixedTrain}$.
According to our hypothesis, we would expect that for relevant concepts we should be able to find neurons with high values in the described metrics, while for non-relevant concepts we should be unable to do so.
For each considered metric, we define a threshold value above which a neuron is considered as a concept neuron for this particular experiment: $0.85$ for Spearman rank-order correlation, $0.95$ for the linear classifier's accuracy, and $0.9$ for the probability density function intersection.

Figure \ref{tab:selected_neurons} shows the amount of identified concept neurons according to each metric for all relevant concepts. For non-relevant concepts, none of the metrics found any concept neurons.
For example, for the concept of $\concept{EmptyTrain}$, we identified $13$ concept neurons based on the Spearman rank-order correlation and $20$ based on the linear classifier's accuracy. In Section \ref{sec:manipulate_perception}, we explore the importance of the selected concept neurons for the overall performance of our method.

Regarding the amount of selected concept neurons, while these are dependent on the specific threshold value, it should be noted that \nna\ contains about $2 \times 10^{6}$ neurons. Thus, as one might expect the amount of selected concept neurons is minuscule (about $0.001\%$) in comparison with the total amount of neurons in a model. Interestingly, we found all selected neurons to be in the dense part of the model, with the more abstract and complex concepts being focused on the later dense layers of the model.

These results seem to confirm our hypothesis, indicating that it is possible to identify neurons in a neural network model whose activations are highly correlated with human-defined concepts, as long as those concepts are relevant for the task being performed by the model.

%% file: tables/selected_neurons_nn_a.tex
\begin{figure}[]
{\center
{\scriptsize	\begin{tabular}{l|c|c|c|}
		\cline{2-4}
		\multicolumn{1}{c|}{}       & Spearman & Accuracy & Intersection \\ \hline
		\multicolumn{1}{|l|}{$\concept{EmptyTrain}$}    & 13       & 20       & 16            \\ \hline
		\multicolumn{1}{|l|}{$\concept{\exists has.PassengerCar}$}  & 15       & 19       & 18           \\ \hline
		\multicolumn{1}{|l|}{$\concept{\exists has.ReinforcedCar}$} & 18       & 22       & 21           \\ \hline
		\multicolumn{1}{|l|}{$\concept{WarTrain}$}      & 13       & 15       & 9            \\ \hline
	\end{tabular}}
	\caption{Amount of concept neurons in \nna\ for each concept.}
	\label{tab:selected_neurons}}
\end{figure}

%% file: manipulating_activations.tex
\subsection{Manipulating a Neural Networks' Perception} \label{sec:manipulate_perception}

We now proceed to test our main hypothesis: that it is possible to manipulate a neural network's perception regarding specific human-defined concepts by manipulating the activations of the concept cells for those concepts.

We test this hypothesis by considering whether the output of a model, for a given sample, that ought to change had a given concept been identified, indeed changes when that concept is injected.
For instance, if we feed an image of a train having a passenger car and no reinforced cars to \nna, which was trained to identify $\concept{TypeA}$ trains, we would expect it to output that this is not a $\concept{TypeA}$ train. However, if we identify the concept neurons for $\concept{\exists has.ReinforcedCar}$ -- having a reinforced car -- and modify their outputs to indicate the presence of a reinforced car, i.e., we \emph{inject} the concept $\concept{\exists has.ReinforcedCar}$, we expect the model to change its output to identifying a $\concept{TypeA}$ train.

To test our hypothesis in the setting of the XTRAINS dataset, we subsample the dataset into four different $1000$ sample sets, for each of the four relevant concepts considered in Section \ref{sec:identify_neurons}, according to the following criteria:
\begin{itemize}
	\item $S1$ - contains $\concept{\neg TypeA}$ samples where the presence of that concept should change \nna's output.
	\item $S2$ - contains $\concept{\neg TypeA}$ samples where the presence of that concept should not change \nna's output.
	\item $S3$ - contains $\concept{TypeA}$ samples where the absence of that concept should change \nna's output.
	\item $S4$ - contains $\concept{TypeA}$ samples where the absence of that concept should not change \nna's output.
\end{itemize}
To determine whether the output of \nna\ should change, we reason with the ontology provided with the dataset.

As the first step in our method is to compute a neurons' sensitivity, we compare the results obtained by the three metrics described in Section \ref{sec:identify_neurons} when applying our method to each of the four described sets. Figure \ref{fig:results_by_affinity}, shows the average number of samples where the \nna's output behaved as expected, when computing the concept neurons' activation values $c_{i}^{\mathcal{M}}$ as the median value of the neurons' activations for both $P_{\concept{C}}$ and $N_{\concept{C}}$, with $|P_{\concept{C}}| = |N_{\concept{C}}| = 1000$.
These results seem to indicate that it is possible to manipulate a neural networks' perception regarding different concepts, by modifying the activations of specific neurons which seem to identify those concepts. This is evidenced by the high percentage of samples where the models' output changed as expected.
Furthermore, these results also indicate that the sensitivity metric based on the intersection of the probability density functions of a neuron for positive and negative samples seems to provide more consistent and better results on average. For this reason, all remaining experiments use this metric to compute the sensitivity of a neuron with regards to a concept ($s_{\concept{C}}$).

\input{tables/affinity_results}

Regarding the method to compute the neurons' activations, $c_{i}^{\mathcal{M}}$, besides using the median value of the neurons' activations as in the previous experiments, we considered two alternatives: computing the mode of the neurons' activations, and computing the values through the method described in \cite{Tucker2021}. Figure \ref{fig:results_by_injection} shows the average results obtained by each method.
Using the median value achieved the best results, having the highest number of correctly classified samples among the three considered metrics for all concepts. We attribute the inferior results obtained by the method in \cite{Tucker2021} to the fact that the information of each concept neuron is quite redundant, and thus the probe used in the method learns to perform its task from just a small subset of those neurons, which might lead to most neurons having their activations unchanged.
In all remaining experiments, we compute $c_{i}^{\mathcal{M}}$ as the median activation value of a neuron. 

\input{tables/injection_results}

Figure \ref{fig:results_by_set} shows the percentage of samples where injecting a concept resulted in the expected \nna\ output for each of the sets described above. Overall the results seem to be quite positive, indicating that typically the model outputs the expected result given the concept being injected.
The result of injecting $\concept{\neg \exists has.PassengerCar}$ in set $S4$ seems to be somewhat inferior to the remaining results, which we believe to be an indication of some spurious correlation learned by \nna.

\input{tables/results_by_set_nna}

The high percentage of samples where the injection of a given sample leads to the expected change in the model's output constitutes strong evidence that it is possible to modify the perception of a neural network regarding specific human-defined concepts, by manipulating the activations of the neurons responsible for the identification of those concepts.

%% file: tables/affinity_results.tex
\begin{figure}
{\center
{\scriptsize
\begin{tabular}{c|c|c|c|}
	\cline{2-4}
	                                 & Spearman & Accuracy       & Intersection   \\ \hline
	\multicolumn{1}{|l|}{$\concept{EmptyTrain}$}    & 98.9\%    & 99.1\%          & \textbf{99.6\%} \\ \hline
	\multicolumn{1}{|l|}{$\concept{\exists has.PassengerCar}$}  & 90.7\%    & \textbf{93.0\%} & 92.9\%          \\ \hline
	\multicolumn{1}{|l|}{$\concept{\exists has.ReinforcedCar}$} & 97.9\%    & \textbf{98.3\%} & 98.2\%          \\ \hline
	\multicolumn{1}{|l|}{$\concept{WarTrain}$}      & 66.6\%    & 65.7\%          & \textbf{99.0\%} \\ \hline
\end{tabular}}
\caption{Correctly classified samples by sensitivity metric.}
\label{fig:results_by_affinity}}
\end{figure}

%% file: tables/injection_results.tex
\begin{figure}
{\center
{\scriptsize
\begin{tabular}{c|c|c|c|}
	\cline{2-4}
	                                 & Median         & Mode  & \cite{Tucker2021} \\ \hline
	\multicolumn{1}{|l|}{$\concept{EmptyTrain}$}    & \textbf{99.6\%} & 99.5\% & 72.4\% \\ \hline
	\multicolumn{1}{|l|}{$\concept{\exists has.PassengerCar}$}  & \textbf{92.9\%} & 92.8\% & 86.5\% \\ \hline
	\multicolumn{1}{|l|}{$\concept{\exists has.ReinforcedCar}$} & \textbf{98.2\%} & 97.7\% & 73.9\% \\ \hline
	\multicolumn{1}{|l|}{$\concept{WarTrain}$}      & \textbf{99.0\%} & 98.6\% & 65.3\% \\ \hline
\end{tabular}}
\caption{Correctly classified samples by method to compute neuron activation.}
\label{fig:results_by_injection}}
\end{figure}

%% file: tables/results_by_set_nna.tex
\begin{figure}
{\center
{\scriptsize
\begin{tabular}{l|l|l|l|l|}
	\cline{2-5}
	\multicolumn{1}{c|}{}            & $S1$    & $S2$                     & $S3$     & $S4$    \\ \hline
	\multicolumn{1}{|l|}{$\concept{EmptyTrain}$}    & 99.7\% & \multicolumn{1}{c|}{-} & 99.5\%  & 99.5\% \\ \hline
	\multicolumn{1}{|l|}{$\concept{\exists has.PassengerCar}$}  & 99.9\% & 98.6\%                  & 98.7\%  & 74.5\% \\ \hline
	\multicolumn{1}{|l|}{$\concept{\exists has.ReinforcedCar}$} & 99.2\% & 98.5\%                  & 96.2\%  & 99.1\% \\ \hline
	\multicolumn{1}{|l|}{$\concept{WarTrain}$}      & 98.4\% & \multicolumn{1}{c|}{-} & 100.0\% & 98.7\% \\ \hline
\end{tabular}}
\caption{Percentage of correctly classified samples in each set.}
\label{fig:results_by_set}}
\end{figure}

%% file: importance_of_neurons.tex
\subsection{Importance of Selected Neurons}

The first two steps in the proposed method have the goal of identifying which neurons in a model act as concept cells for a given concept of interest. These neurons are then used to manipulate a neural networks' perception regarding that concept. Thus, one might wonder about how important the specific set of selected concept neurons is to the overall performance of the method. This Section addresses the effects of varying the threshold value of function $s_{\concept{C}}$.

\begin{figure}
	\includegraphics[width=0.98\linewidth]{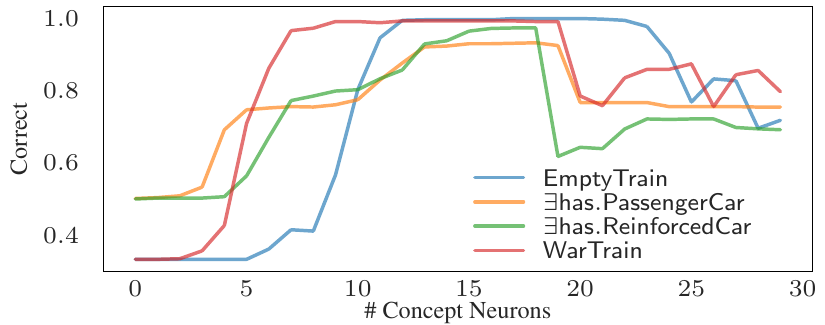}
	\caption{Correct samples by amount of selected concept neurons.}
	\label{fig:performance_by_neurons}
\end{figure}

In order to understand how dependent the methods' performance is on the amount of concept neurons, we measure, for each concept, the ratio of samples where the output of the model is consistent with the manipulation performed. Figure \ref{fig:performance_by_neurons}, shows the average result obtained for all sets described in Section \ref{sec:manipulate_perception} for each amount of considered concept neurons. 

Our results indicate that if function $s_{\concept{C}}$ is either too restrictive, excluding many neurons which act as concept cells, then the few considered ones might be unable to affect the model and consistently produce the desired effects in the model -- this is shown by the significant increase in performance as more concept neurons are initially added. However, if function $s_{\concept{C}}$ too tolerant and starts to include neurons that are not effectively concept cells for a given concept performance starts to degrade -- as shown by the sharp decrease in performance for higher number of considered concept neurons. 

As expected, the performance of the proposed method is heavily dictated by the set of selected concept neurons. These results shows how important it is to adjust the threshold value of function $s_{\concept{C}}$ for each concept of interest, which might be done using a simple grid search with cross-validation. 

%% file: methods_cost.tex
\subsection{Counterfactual's Cost} \label{sec:cost}

The results so far suggest that it is possible to manipulate a neural network's perception of specific human-defined concepts by modifying the activation of neurons which seem to be identifying those concepts. 
However, they have assumed that there exists plenty of labeled data available, using a total of $2000$ samples to compute neurons' sensitivity values and the activation values to be injected for each concept.
In this section, we assess whether the proposed method is practical when taking into account the amount of necessary data.

To assess the cost of applying this method, and how the amount of available labeled data impacts our capacity to properly modify a networks' perception, we compare the average results obtained in the $4$ sample sets defined in Section \ref{sec:manipulate_perception}, for each injected concept, when using different amounts of available data.
We keep the amount of samples used to validate the threshold value of $s_{\concept{C}}$ unchanged throughout this comparison, using $100$ samples as validation.

\begin{figure}
	\includegraphics[width=0.98\linewidth]{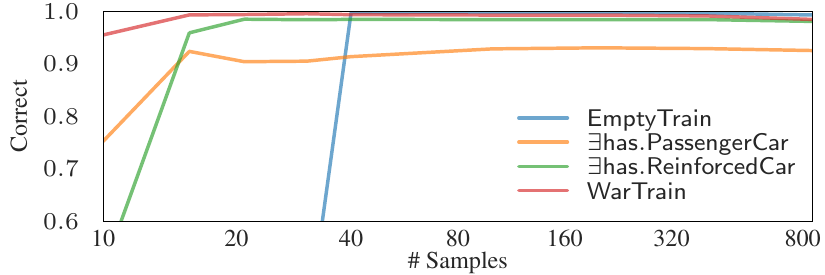}
	\caption{Correct samples by amount of available labeled data.}
	\label{fig:performance_by_data}
\end{figure}

Figure \ref{fig:performance_by_data}, shows the average ratio of correctly classified samples after injection of a given concept, for each considered concept. It is observable that even with as few as $40$ labeled samples -- $20$ samples where the concept is present and $20$ where it is absent -- it is possible to manipulate the neural networks perception for each considered concept with a high degree of success.

The method's performance starts to significantly drop when the amount of labeled data is insufficient to be able to properly identify the concept neurons for a given concept. This seems to indicate that as long as there is enough labeled data to identify which neurons in a model are sensitive to the concepts we want to inject, we should be able to manipulate a neural network's perception with regards to that concept.

%% file: generating_counterfactuals.tex
\section{Interpreting Neural Networks} \label{sec:interpret_nns}

So far, we described a method to generate counterfactuals for a neural network model by modifying the activations of neurons which act as concept cells for human-defined concepts. 
Through this method, it is possible to understand how different concepts influence the output of a model.
What if one wants to be able to understand how a model relates different concepts which are not represented in the model's output?

For example, one might be interested in understanding whether \nna\ has learned that $\concept{EmptyTrain}$s do not have any passenger cars, and that if $\concept{\exists has.PassengerCar}$ then that train is not an $\concept{EmptyTrain}$. Although both of these concepts are not represented in \nna's output they are both relevant for its task and one might be interested in assuring that the model is capable of understanding the relationship between both.

In order to understand whether a given concept which is not represented in the output of a model is being identified by it, we make use of \emph{mapping networks} \cite{SousaRibeiro2021}. These are small neural networks trained to identify in the activations of a neural network model whether a given human-defined concept was identified.

To verify whether \nna\ has learned the described relations between both concepts, we train two mapping networks for the concepts of $\concept{EmptyTrain}$ and $\concept{\exists has.PassengerCar}$ -- both achieving an accuracy of more than $99\%$ on a $1000$ balanced test set. To test whether \nna\ has learned that empty trains do not have any passenger cars, we inject the concept $\concept{EmptyTrain}$ on $1000$ samples of trains with passenger cars. Through the mapping network for $\concept{\exists has.PassengerCar}$, we verify that in $95.6\%$ of the samples, after injection, the concept $\concept{\exists has.PassengerCar}$ goes from being present to being absent. This is strong indication that this model has learned that empty trains do not have passenger cars.

Similarly, we test whether \nna\ has learned that if a train has a passenger car, then it is not an empty train. We inject the concept $\concept{\exists has.PassengerCar}$ on $1000$ samples of empty trains, and observe that in only $2\%$ of them the concept $\concept{EmptyTrain}$ turns absent. This indicates that \nna\ has not learned that if a train has a passenger car, it is not an empty train.

This example illustrates a possible application of our method, namely to investigate whether certain relations between user-defined concepts are encoded in the model.
But more importantly, the model seems to naturally integrate the injected information, impacting the related concepts, as if the injected concept was truly being perceived by the model.

%% file: correcting_errors.tex
\section{Correcting a Neural Network's Misunderstandings} \label{sec:correct_nns}

When a neural network model provides an incorrect result, it is difficult to interpret the cause of the error.
It might often be because the model was unable to correctly perceive part of what is represented in its input, in which case the proposed method might be used to ``correct'' a model's wrong results.

To test our hypothesis, we select all XTRAINS samples where \nna\ provides an incorrect result. We train a mapping networks for the concepts $\concept{\exists has.ReinforcedCar}$ and $\concept{\exists has.PassengerCar}$ -- with an accuracy of more than $99\%$ on a $1000$ balanced test set -- and use them to identify whether \nna\ provided a wrong result because either of these concepts were not perceived by the model when they should have.

We select all samples where \nna\ provides a false negative result, indicating that a sample input was not of $\concept{TypeA}$ when it was. From these samples, we observe that in those where the concept of $\concept{\exists has.ReinforcedCar}$ was not identified, but present in the input, injecting $\concept{\exists has.ReinforcedCar}$ led to the output being corrected in $96.1\%$ of the samples.
Similarly, when considering the samples where $\concept{\exists has.PassengerCar}$ was not identified, but present in the input, injecting it led to the output being corrected in $98.7\%$ of the samples.

These results provide evidence that our method is applicable even when a model provides incorrect results, allowing one to test whether different concepts might be the responsible for the provided result. This enables users to further understand why their models achieved an incorrect output, and identify potential flaws in the model or in its training.

%% file: validation_realworld_data.tex
\section{Validation with Real World Data} \label{sec:validation}

So far, we have only considered a synthetic dataset. We now consider whether we can replicate similar results with real data.
To perform this validation, we consider the setting of the ImageNet dataset \cite{Russakovsky2015}, and examine the pre-trained MobileNetV2 model from \cite{Sandler2018}.

The proposed method is based on the assumption that by identifying which neurons in a model are responsible for identifying a given human-defined concept, we are able to modify the model's perception of that concept by modifying those neuron's activations. Thus, we focus our validation on whether we can identify the neurons responsible for a given human-defined concept, and whether we are able to modify a model's outputs by injecting that concept.

\begin{figure}
	\hfill
	\hfill
	\includegraphics[width=0.1\textwidth]{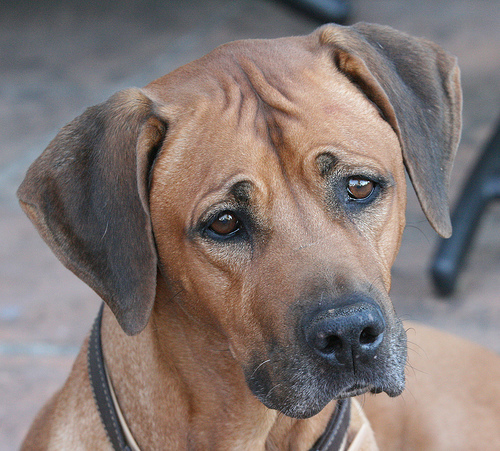}
	\hfill
	\includegraphics[width=0.1\textwidth]{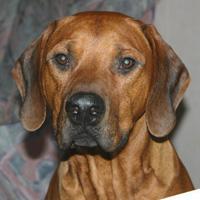}	
	\hfill
	\includegraphics[width=0.1\textwidth]{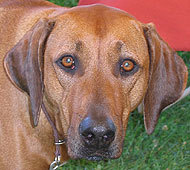}	
	\hfill
	\hfill
	\caption{Images of `Rhodesian ridgeback dog face'.}
	\label{fig:dog_face}
\end{figure}

In order to test our method in this setting, we selected a random output class: `Rhodesian ridgeback', which is a specific dog's breed, and define the concept of `Rhodesian ridgeback dog face' by selecting a set of $98$ images from ImagenetNet where a Rhodesian ridgeback dog is observable and its face is centered and completely visible, as illustrated in the samples shown in Figure \ref{fig:dog_face}.
We select this concept based on the assumption that it is a useful concept for the model to classify images of Rhodesian ridgeback dogs.

To test whether by injecting this concept we were able to modify the model's perception, we used all $329$ samples -- from training and validation sets -- where the model outputs a false negative for the Rhodesian ridgeback class considering its top-1 result. The injection of the concept of `Rhodesian ridgeback dog face' led $48\%$ of these samples to output the class of Rhodesian ridgeback.
If we consider the $38$ samples where the model outputs a false negative considering its top-5 outputs, we were able to modify the model's output in $71.1\%$ to Rhodesian ridgeback class.

Since the concept of `Rhodesian ridgeback dog face' is not sufficient by itself to lead the model to output the class of `Rhodesian ridgeback', these results constitute evidence that it is being used by the model, and that by manipulating its concept neurons we are able to modify the model's output.

\begin{figure}
	\hfill
	\hfill
	\includegraphics[width=0.12\textwidth]{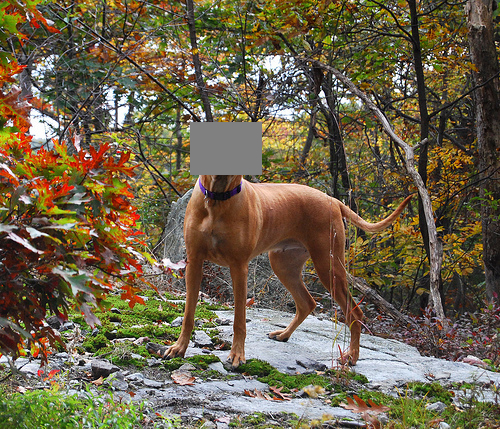}
	\hfill
	\includegraphics[width=0.12\textwidth]{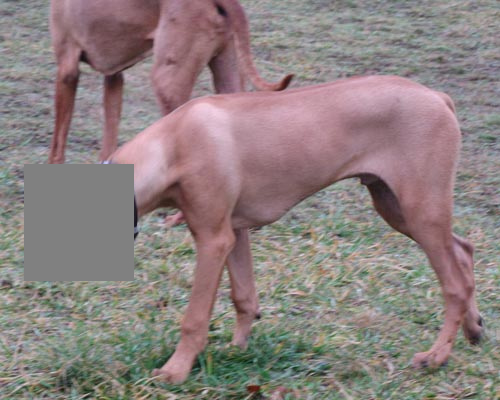}	
	\hfill
	\includegraphics[width=0.12\textwidth]{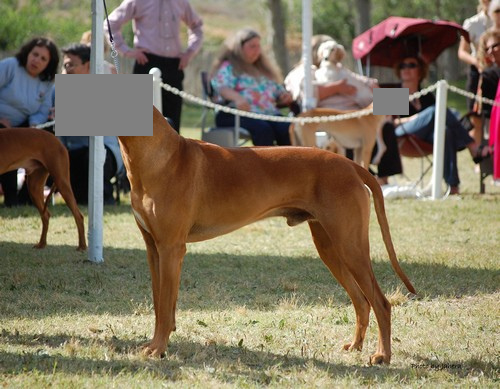}	
	\hfill
	\hfill
	\caption{Images of censored `Rhodesian ridgeback dog face'.}
	\label{fig:censored_dog_face}
\end{figure}

We further test the assumptions that the concept of `Rhodesian ridgeback dog face' is important for the model to be able to recognize that breed, and that we are able to generate counterfactuals by injecting that concept, by first censoring every `Rhodesian ridgeback dog face' in ImageNet's validation set, as shown in Figure \ref{fig:censored_dog_face}.
We then measure the accuracy of the model before censoring the dog's faces, after censoring them, and after injecting the concept of `Rhodesian ridgeback dog face' on the censored ones. The results, shown in Figure \ref{fig:dog_face_exp}, indicate that censoring the dog's face leads to a significant drop in accuracy, which seems to be evidence of its importance for the model. Then, when the concept is injected even after the faces are censored, we observe a big increase in accuracy, indicating that the injected concept was successful in providing some of the information removed by the censoring and helping the model provide the correct classification.

\input{tables/dogface_table}

These results show us that even in a setting with real data, a moderately sized model ($\approx 7 \times 10^6$ neurons), and few labeled data, it is possible to identify which neurons in a model act as concept cells for a given concept and, more importantly, manipulate the neural network's perception of that concept by manipulating the activations of those neurons.

%% file: tables/dogface_table.tex
\begin{figure}
{\center
{\scriptsize
\begin{tabular}{c|c|c|c|}
	\cline{2-4}
	\multicolumn{1}{l|}{}       & \multicolumn{1}{l|}{Not Censored} & \multicolumn{1}{l|}{Censored} & \multicolumn{1}{l|}{Censored + Injection} \\ \hline
	\multicolumn{1}{|c|}{Top-1} & 70\%                              & 36\%                          & 64\%                                      \\ \hline
	\multicolumn{1}{|c|}{Top-5} & 100\%                             & 64\%                          & 88\%                                      \\ \hline
\end{tabular}}
	\caption{Model accuracy regarding `Rhodesian ridgeback' class.}
	\label{fig:dog_face_exp}}
\end{figure}

%% file: related_work.tex
\section{Related Work}
\label{sec:related_work}

The last few years has seen an increase in the development of methods for interpreting artificial neural networks, with proxy-based methods being one of the most popular approaches (c.f., \cite{Gilpin2018}). These methods aim to substitute the model being explained for one that is inherently interpretable and which exhibits similar behavior.  
In contrast, our approach to generate counterfactual scenarios does not require changing or substituting the original model, which might not always be feasible.

Another popular approach is that of saliency and attribution methods (c.f., \cite{Li2021}), where a model's behavior is explained by attributing a contribution value to each input feature representing its contribution to a given prediction.
Although these methods might provide some insights into which features contributed most for a given prediction, they do not provide clarification regarding why those contributions justify the output, leaving the burden of understanding how those contributions are related with the specific output to the user.

Some counterfactual-based methods (c.f., \cite{Guidotti2022}) suffer from similar issues, providing a counterfactual sample but lacking clarification of why that sample leads to a different result.
We believe that counterfactual methods would benefit from abstracting away from only generating specific counterfactual samples, to describing in a human-understandable way why those counterfactuals lead to some other particular output based on how they affect the model. 

Recently, neuro-symbolic approaches have aimed at understanding whether models are sensitive to specific human-defined concepts \cite{Kim2018}, and how we can leverage the representations of those concepts in a model to provide explanations for its outputs \cite{SousaRibeiro2021}, as well as how to produce human-understandable theories representing the classification process of a model \cite{Ferreira2022}.
In this paper, we link both approaches through a method that focuses on what a model is perceiving and enables the generation of counterfactuals via manipulation of what is being perceived wrt. human-defined concepts.

%% file: conclusion.tex
\section{Conclusions} \label{sec:conclusions}

In this paper, we proposed a method to generate counterfactuals regarding what a neural network model is perceiving about human-defined concepts, enabling users to inquire the model, and understand how its outputs are dependent on those concepts.
To this end, we explored how to identify which neurons in a model are sensitive to a given concept, and how to leverage such neurons to manipulate a model into perceiving that concept.
Through experimental evaluation, we show that it is possible manipulate a neural network's perception regarding different concepts, requiring few labeled data to do so, and without needing to change or retrain the original model.

We provide a formalization of the proposed method, and illustrate how it can be applied to generate counterfactuals, but also how to use the it to better understand how different concepts are related within a model, and how to inspect and correct a model's misclassifications.

We conclude that for concepts that are related to the task of a neural network, it is often the case that there exist a few specific neurons which encode that concept and are responsible for its identification within the model. Modifying the activations of these neurons, similarly to stimulating concept cells in the human brain, allows one to \emph{trick} the model that it is perceiving that concept. This simple method seems effective in allowing sophisticated manipulations of high-level abstract concepts, enabling users to explore, with little effort, how the model would respond in different scenarios.

In the future, we are interested in exploring how to leverage this method to search for learned biases, and identify missing or undesired associations between concepts.

\section*{Acknowledgments}
The authors would like to acknowledge the support provided by FCT through PhD grant (UI/BD/151266/2021) and strategic project NOVA LINCS (UIDB/04516/2020).